\documentclass[letterpaper, 10 pt, conference]{ieeeconf}  

\IEEEoverridecommandlockouts                              

\usepackage{cite}
\usepackage{amsmath,amssymb,amsfonts}
\usepackage{algorithmic}
\usepackage{graphicx}
\usepackage{textcomp}
\usepackage{xcolor}

\usepackage[nolist]{acronym}
\usepackage[colorinlistoftodos, textsize=footnotesize, textwidth=20mm]{todonotes}
\usepackage{svg} 
\usepackage{siunitx} 

\usepackage{tabularx}
\usepackage{tikz}

\newcommand\circlearound[1]{%
  \tikz[baseline]\node[draw,shape=circle,anchor=base,scale=0.70] {#1} ;}

\def\BibTeX{{\rm B\kern-.05em{\sc i\kern-.025em b}\kern-.08em
    T\kern-.1667em\lower.7ex\hbox{E}\kern-.125emX}}

\title{\LARGE \bf
Should Teleoperation Be like Driving in a Car? \\
Comparison of Teleoperation HMIs
}

\author{Maria-Magdalena Wolf$^{*,1}$, Richard Taupitz$^{*}$, Frank Diermeyer$^{*}$
\thanks{The research was funded by the Federal Ministry of Economic Affairs and Climate Action of Germany (BMWK) within the project SAFESTREAM.}
\thanks{$^{*}$The authors are with the Institute of Automotive Technology, Technical University of Munich (TUM), Boltzmannstr. 15, DE-85748 Garching
bei M\"unchen, Germany}%
\thanks{$^{1}$Corresponding author:
        {\tt\small maria.wolf@tum.de}}%
}

\begin{document}

\maketitle
\thispagestyle{empty}
\pagestyle{empty}


\begin{abstract}

Since Automated Driving Systems are not expected to operate flawlessly, Automated Vehicles will require human assistance in certain situations.
For this reason, teleoperation 
offers the opportunity for a human to be remotely connected to the vehicle and assist it.
The Remote Operator can provide extensive support by directly controlling the vehicle, eliminating the need for Automated Driving functions.
However, due to the physical disconnection to the vehicle, monitoring and controlling is challenging compared to driving in the vehicle.
Therefore, this work follows the approach of simplifying the task for the Remote Operator by separating the path and velocity input.
In a study using a miniature vehicle, different operator-vehicle interactions and input devices were compared based on collisions, task completion time, usability and workload.
The evaluation revealed significant differences between the three implemented prototypes using a steering wheel, mouse and keyboard or a touchscreen.
The separate input of path and velocity via mouse and keyboard or touchscreen is preferred but is slower compared to parallel input via steering wheel.

\end{abstract}




\begin{acronym}

    \acro{ro}[RO]{Remote Operator}
    \acroplural{ro}[ROs]{Remote Operators}

    \acro{av}[AV]{Automated Vehicle}
    \acroplural{av}[AVs]{Automated Vehicles}

    \acro{ads}[ADS]{Automated Driving System}

    \acro{gui}[GUI]{Graphical User Interface}
    \acroplural{gui}[GUIs]{Graphical User Interfaces}
    
    \acro{hmi}[HMI]{Human-Machine Interaction}
    \acroplural{hmi}[HMIs]{Human-Machine Interactions}

    \acro{kpi}[KPI]{Key Performance Indicator}
    \acroplural{kpi}[KPIs]{Key Performance Indicators}
    
    \acro{mk}[M\&K]{Mouse {\&} Keyboard}

    \acro{sus}[SUS]{System Usability Scale}
    
    \acro{sw}[SW]{Steering Wheel}

    \acro{tod}[ToD]{Teleoperated Driving}

    \acro{ts}[TS]{Touchscreen}

    \acro{ui}[UI]{User Interface}
    \acroplural{ui}[UIs]{User Interfaces}

\end{acronym}


\section{Introduction}

During teleoperation, the driver no longer needs to be on site in the vehicle but can support and control the vehicle remotely. Teleoperation holds the promise of greater flexibility for the \ac{ro} and increased efficiency, while maintaining the same level of safety for the vehicle occupants and other road users.
Furthermore, teleoperation can be used to support \acp{av} in the event of failures or uncertainties in the \ac{ads}.

Several approaches of teleoperation concepts to remotely support \acp{av} have been reported in the literature \cite{Maj2022}. Among these, the most widely researched and implemented teleoperation concept is Direct Control \cite{Ama2022}.
In Direct Control, no automation of the vehicle is required and control signals, such as steering angle and velocity, are sent directly to the vehicle and executed.
Start-ups, such as Fernride and Vay, have already established their first business models with Direct Control.
However, Direct Control is demanding for the \acp{ro} as they have to fulfill the entire driving task remotely and physically disconnected from the vehicle \cite{Ten2022}.
Thus, the \ac{ro} provides longitudinal and lateral guidance through continuous input, e.g., via a steering wheel and pedals, under the challenges of reduced situational awareness \cite{Mut2021} and transmission latency \cite{Geo2020}.
The resulting high workload leads to an increased probability of \ac{ro} errors and potential accidents \cite{Une2020}.
This makes it particularly important to reduce the \ac{ro} workload during teleoperation by improving the \ac{hmi}.
However, previous approaches, such as displaying the predicted vehicle position to illustrate the latency \cite{Chu2016} or various visualization concepts \cite{Geo2020b}, including the use of a head-mounted display \cite{Hos2016}, only consider the output to the \ac{ro} but not the \ac{ro} input.
Another way to solve the remote driving task and simplify the interaction between the \ac{ro} and the vehicle is to execute the path and velocity guidance sequentially.
Consequently, the \ac{ro} first specifies the route to be followed
and then sets the velocity while the vehicle travels along the predefined route.
This division of tasks suggests a reduction in the \ac{ro} workload while appearing to result in longer execution times.

Nevertheless, the division of tasks has further advantages, as the vehicle can follow the predefined path to a standstill if the connection is lost. In addition, the \ac{ro} could be further relieved by the \ac{av} automatically traveling along the predefined path.
This type of interaction is already provided for teleoperation concepts, such as Trajectory Guidance or Waypoint Guidance \cite{Maj2022}.
As vehicle automation will continue to advance, it is likely that the \ac{ro} will only have to assist the \ac{av} instead of steering it. Therefore it can be assumed that input devices, such as the steering wheel, will no longer be needed.
In order to design a standardized workplace for occasionally remote driving and remote assistance, it is necessary to investigate, which input devices are suitable for both approaches.

The presented work aims to improve the \ac{hmi} in teleoperation by deriving design suggestions for the \ac{ro} user interface
to simplify the remote driving task.
Therefore, we make the following contributions:
\begin{itemize}
    \item Development of a novel concept for separate path and velocity input
    \item Comparison of three \acp{hmi} with different input devices in an expert study performed with a miniature vehicle
    \item Evaluation of collisions, time, usability and workload
    \item Derivation of design suggestions
\end{itemize}


\section{Related Work}

\subsection{Taxonomy}

\begin{figure}
  \centering
  \includegraphics[width = 8.7cm]{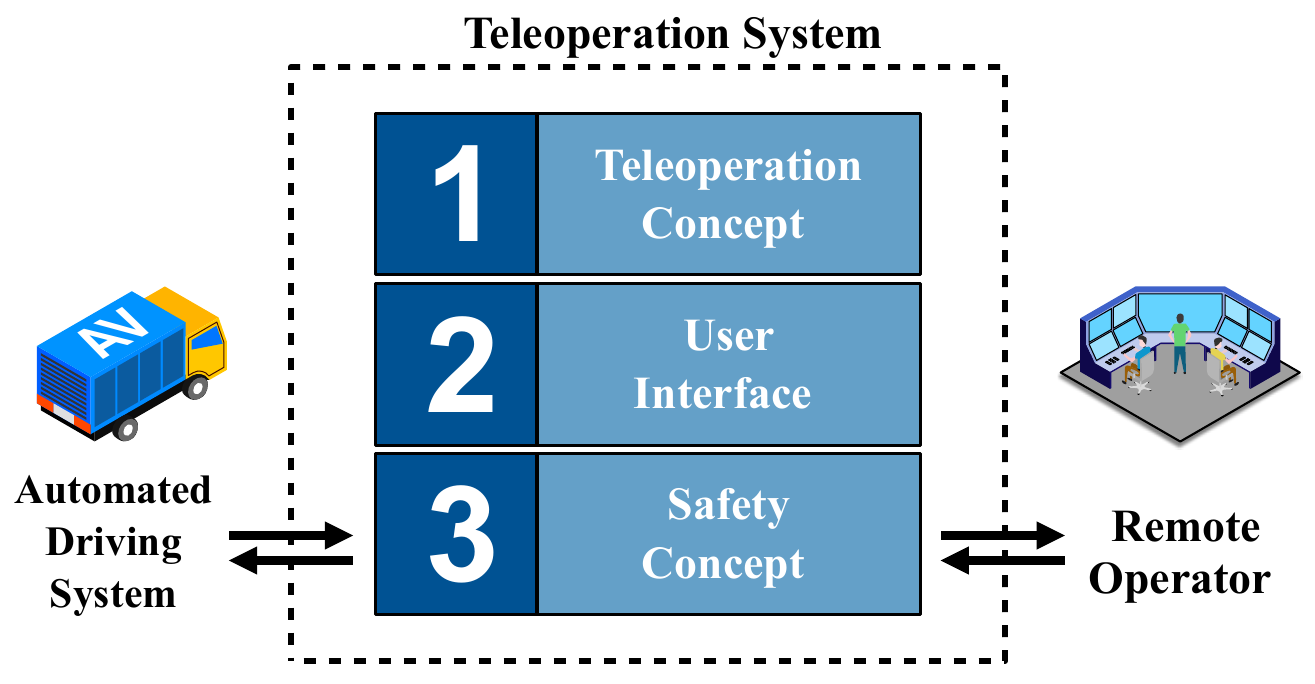}
  \caption{A teleoperation system comprises a teleoperation concept, a user interface and a safety concept}
  \label{fig:teleopSystem}
\end{figure}

A teleoperation system comprises three components \cite{Brecht2024}, as shown in Fig.~\ref{fig:teleopSystem}:
\begin{itemize}
    \item the \textit{Teleoperation Concept}, that defines the general interaction between the \ac{ro} and vehicle, 
    \item the \textit{\ac{ui}}, through which the \ac{ro} and system communicate with each other and 
    \item a \textit{Safety Concept}, which supports the execution with protective measures.
\end{itemize}
In this work, the \ac{ui} is analyzed in more detail. 
To enable communication between the \ac{ro} and the teleoperation system, the outputs of the system must be transmitted to the \ac{ro} and the \ac{ro} inputs must be transferred to the system.

Regarding the output to the \ac{ro}, humans can only perceive information via a limited number of senses \cite{Mller2016}. 
In the case of teleoperation, visual, auditory, haptic or vestibular transmission is suitable \cite{Zhao2023}.
Thereby, it should be noted that people mainly perceive their surroundings visually \cite{Colavita1974}.
Therefore, an output device, such as one or more screens or a head-mounted display, is needed. 
Nevertheless, the display is not trivial, requiring a precise selection of information, as well as presentation style, position or size. 

In addition, there are a variety of input modalities that allow the \ac{ro} to pass on commands to the system.
The most conventional input is haptic, whereby input via gesture or voice control is conceivable \cite{Chen2007}.
Typical haptic input devices for teleoperation are the \ac{sw} and pedals, a \ac{ts}, \ac{mk} or a joystick.
The challenge remains in finding the right input and output setup from the multitude of possibilities.

\subsection{Teleoperation User Interface}

According to surveys, the \ac{ui} in teleoperation has been studied primarily in the context of remote driving \cite{Maj2022, Ama2022}.
In addition to studies on the multi-modal transmission of information to improve situational awareness and the immersion of the \ac{ro} \cite{Zhao2023, Chen2007, Hos2016b, Kallioniemi2021}, research focuses on the visual output and thus the \ac{gui}.

Previous publications explore the visualization of latency in order to ensure a safe journey even in the worst-case scenario of a connection loss. In this regard, Tang et al.~\cite{Tan2014} introduced the Free Corridor to reach a safe end state. Another example to handle latency issues is the visualization of the predicted vehicle position \cite{Chu2016}.
Moreover, there are various investigations into how the video should be displayed with regard to the necessary video quality \cite{Geo2020b}, the point of view \cite{Boker2023} and field of view \cite{Voysys2020}. 
Beyond that, other publications deal with a diverse set of display concepts \cite{Geo2020b, Cab2019, Geo2019} and output devices like a head-mounted display 
\cite{Hos2016, She2016, Bou2017, Geo2020b} to accommodate all necessary information on display surfaces.

While individual elements of the \ac{gui} have often been examined, only a few holistic \ac{gui} designs and evaluations have been performed. Among these, Gafert et al.~\cite{Gafert2023} present a \ac{gui} for Direct Control and compare the wealth as well as the relevance of information in the orientation and navigation phase in a real driving study with a test vehicle and a head-mounted display.
Another holistic \ac{gui} consisting of more than 120 interactive screens was developed by Tener and Lanir~\cite{Tener2023b}. Although the prototype has not yet been evaluated, it is one of the first to incorporate the concepts of remote assistance, whereby the \ac{ro} supports the vehicle not by sending control signals but by providing planning decision. The developed \ac{gui} combines various teleoperation concepts, such as selecting \ac{av}-proposed options, classifying unrecognized objects and entering paths \cite{Tener2023}.
However, the \ac{gui} is currently only available as a click-dummy, not as a real implementation. Evaluated results are also pending.

Therefore, the following section takes a dedicated look at the investigation of the interaction of the separate path and velocity specification, which will be considered in the course of this work.

\subsection{Separate Path and Velocity Specification}

In industry, the concept of path specification by the~\ac{ro} stands out, where the \ac{av} automatically follows the generated path without requiring velocity input from the \ac{ro}. This teleoperation concept is known as Waypoint Guidance \cite{Maj2022}.

Cruise \cite{Cruise2021}, Motional \cite{Motional2022, Motional2023} and Zoox \cite{Zoox2020} showcase initial approaches and \acp{gui}. In all these introduced \acp{gui}, the \ac{ro} is shown several video streams and an abstract representation of the surroundings, including the road layout and recognized objects. Motional \cite{Motional2022} additionally incorporates traffic light visualization. 
In order to understand the vehicle's mission and problem, the actual trajectory of the \ac{av} to be driven as well as blocking objects or the blocked section of the trajectory are color-coded.
The \ac{ro} task is to provide guidance by setting an alternative path via waypoints in the 2D representation.
Input is made either via mouse clicks \cite{Motional2022b} or a \ac{ts} \cite{Zoox2020}, whereby the \ac{av} recalculates its trajectory based on the suggested path and drives along automatically reacting to dynamic obstacles.
Additionally, the \ac{ro} can set conditions for the vehicle to follow the path, such as waiting for nearby objects to clear or for traffic lights to turn green \cite{Cruise2021}.

Initial research into control via waypoints sent to a real vehicle was conducted by Kay \cite{Kay1997} in a study with 19~participants.
Vehicle velocity was controlled by a safety driver in the vehicle.
The aim of the study was to investigate the functionality and performance of the waypoint specification under the aspect of latency and limited image data by measuring \acp{kpi} like retries, total time or mean number of sent points. Effort in improving the \ac{ro} workload is missing.

According to the published video \cite{Schitz2021b}, the interactive corridor-based path planning by Schitz et al. \cite{Schitz2021} is also based on the \ac{ro} specifying waypoints with mouse clicks, thus creating a corridor for potential routes. 
Within this corridor, the \ac{av} calculates a collision-free trajectory.
Although the technical functionality and real-time capability was demonstrated simulatively, as well as in a vehicle implementation, no further evaluation was carried out.

In the concept developed in the 5GCroCo project \cite{5gcroco2021}, the \ac{ro} sets waypoints on the video stream, which creates path segments that are then validated by the \ac{av}, released by the \ac{ro} and finally traveled by the \ac{av}.
\ac{ro} inputs are made via \ac{mk}. The \acp{kpi} evaluated as part of the project purely relate to data transmission.

Analogue to the 5GCroCo project \cite{5gcroco2021}, in the Trajectory Guidance concept developed by Majstorovic et al. \cite{Maj2024} 
the \ac{ro} clicks waypoints creating segments that are checked by the \ac{av}, approved by the \ac{ro} and executed by the \ac{av} at a fixed maximum speed. 
In contrast to the 5GCroCo approach \cite{5gcroco2021}, the waypoints are clicked on a separate lane-let map and the video images are arranged cylindrically around the vehicle model, allowing the selection of a first-person perspective.
Although both Trajectory Guidance and Direct Control were implemented on a test vehicle, no comparative study with test subjects was performed.

In addition to the technical concepts, Kettwich et al. \cite{Kettwich2021} designed a \ac{gui} for an \ac{ro} workstation on six screens and an additional \ac{ts} for Waypoint Guidance \cite{Maj2022}. Three screens in the top row show video streams of the vehicle's surroundings. The other three screens below visualize an overview map, detailed vehicle information and the current disturbances. Moreover, the \ac{ts} shows a bird's eye view map of the vehicle's immediate surroundings including the originally planned trajectory where the \ac{ro} sets waypoints for path specification \cite{DLR2022}. It is assumed that the \ac{av} travels the route at its own velocity.
The \ac{gui} has so far been tested as a click-dummy without a driving task and only with offline videos. 
The study (N=13) achieved low perceived workload, medium situation awareness but high usability and user acceptance.

Besides waypoint entry, there are other options to input path and velocity separately.
Gnatzig et al.\cite{Gna2012} introduce an approach in which the \ac{ro} specifies trajectory segments via \ac{sw} and accelerator pedal. The curvature of the segment is determined by the steering angle, whereas the length of the trajectory sets the velocity adapted by the pedal. After the \ac{ro} release, the trajectory segment is sent to the \ac{av}.
A test drive with a real vehicle demonstrated no immediate reduction in completion time, but improves safety in the case of connection loss or high latency compared to Direct Control approaches.

Another suggestion for entering the path is walking in a large-scale haptic interface \cite{Fen2021}. However, the realization of a physical representation of the vehicle environment seems impossible, so the approach will not be discussed further.

In summary, for separate path and velocity input the path is mainly defined via waypoints, whereby in most cases the velocity no longer has to be specified by the \ac{ro} but is selected automatically by the \ac{av}. 
The work published to date comprised either technical implementations without sufficient evaluation in studies, or click-dummy evaluations without a technical implementation to a vehicle.
Thus, there remains a need for investigations in human subject studies with real vehicle implementations.
Moreover, no comparative studies have yet been published to compare the results with the widely used Direct Control concept.
Therefore, this work deals with the comparison of separate path and velocity input with Direct Control using different input devices.


\section{Methodology}

\begin{figure}[htbp]
  \centering
  \includegraphics[width = 8.7cm]{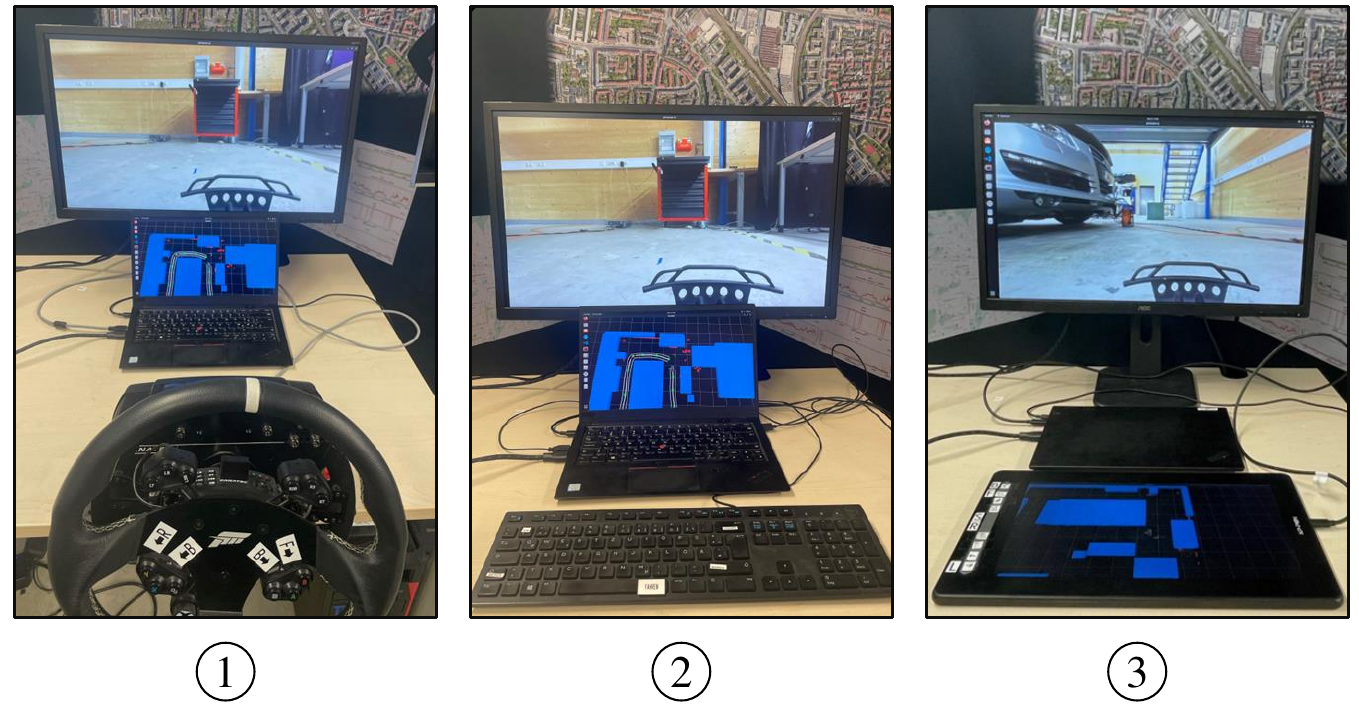}
  \caption{Three implemented \ac{hmi} variants with the display of camera image and map on separate monitors: \protect\circlearound{1} Parallel path and velocity input using a steering wheel, \protect\circlearound{2} Path input via mouse clicks on map and velocity input using keys, \protect\circlearound{3} Path input via stylus clicks on touchscreen map and velocity input using physical touchscreen buttons
\label{HMI_Konzepte}
}
\end{figure}

In this work, an \ac{hmi} for separate input of path and velocity was developed and implemented using two different input devices to compare it with conventional Direct Control in an expert study (N=7).
Therefore, three prototypes for teleoperation
(Table \ref{tab1} and Fig.~\ref{HMI_Konzepte}) were evaluated on an approximately 30\;meter long test track with a miniature vehicle to derive initial trends in the differences and suggestions for improvement regarding the new operator-vehicle interaction.

\begin{table}[htbp]
\caption{Compared HMI Variants}
\begin{center}
\vspace{-0.5cm}
\begin{tabular}{|l|l|l|}
\cline{1-3}
Input & \hspace{-1mm}Path Input Device & \hspace{-1mm}Velocity Input Device\\
\cline{1-3}
\hspace{-1mm}\circlearound{1} Parallel & \hspace{-1mm}Steering Wheel & \hspace{-1mm}Buttons on Steering Wheel\\
\hspace{-1mm}\circlearound{2} Sequential & \hspace{-1mm}Mouse & \hspace{-1mm}Keys on separate Keyboard\\
\hspace{-1mm}\circlearound{3} Sequential & \hspace{-1mm}Stylus on Touchscreen\hspace*{-0.5mm} & \hspace{-1mm}Phys. Buttons on Touchscreen\hspace*{-0.5mm}\\
\cline{1-3}
\end{tabular}
\label{tab1}
\end{center}
\end{table}

\vspace{-0.8cm}

\subsection{Design}

\begin{figure}
  \centering
  \includegraphics[width = 8.7cm]{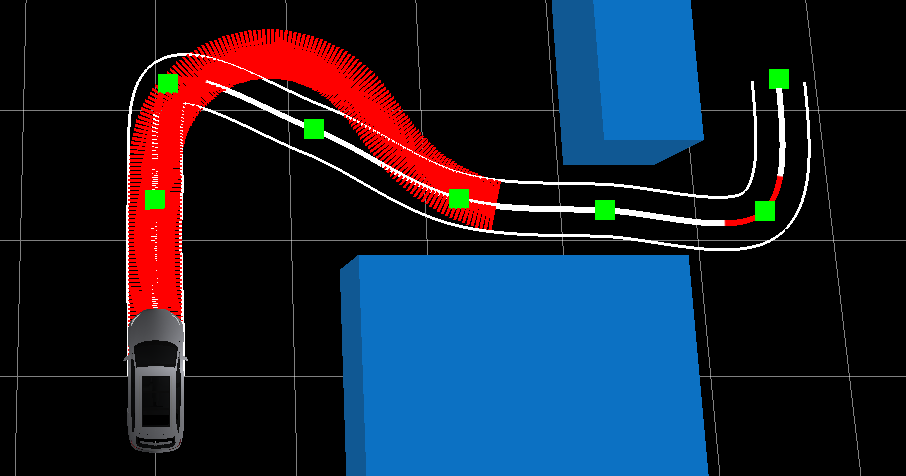}
  \caption{
  The set waypoints are visualized as green squares and connected by a center line with two parallel lines marking the vehicle's width. The red center line indicates impassable curve radii, with the red path simulating the actual vehicle behavior.}
  \label{fig:Design}
\end{figure}

To decide on an \ac{hmi} design for separate path and velocity input, functionalities were collected in a morphological box and discussed with experts from the Institute of Automotive Technology at the Technical University of Munich.
The discussion resulted in an \ac{hmi} design that contains the components shown in Fig.~\ref{fig:Design} and the attached video link\footnote[1]{Video of the \ac{hmi}: https://youtu.be/{\_}e8sBJqupbg}.

To enable the \ac{ro} to set a valid path, the vehicle’s surroundings are represented on two screens showing a camera image and a map where the waypoints are set. Thereby, the monitor and laptop only serve as display devices.
First, the \ac{ro} specifies individual waypoints on the map by clicking the mouse in \ac{hmi} variant \circlearound{2} or using the stylus on the touchscreen in \circlearound{3}. The waypoints are automatically combined in the clicked sequence to form a driving path. During the interaction, a new waypoint is already selected, so the \ac{ro} only has to set the point with one click. Furthermore, the user can edit the path by clicking, holding and moving an existing waypoint via mouse \circlearound{2} or stylus on the touchscreen \circlearound{3}. Deleting a point at the end of the path is possible using a defined key on the keyboard \circlearound{2} or a button on the touchscreen \circlearound{3}. To visualize to the \ac{ro} whether the specified path is compatible with the vehicle’s kinematics, the center line is red-colored as soon as the curvature is outside the maximum steering angle of the vehicle. In order to understand the vehicle behavior generated by the \ac{ro} input, the user can display the route calculated by the vehicle in red by using a defined key \circlearound{2} or a button \circlearound{3}.
After entering the path, the \ac{ro} can set the vehicle in motion by accelerating or decelerating it using defined keys on the keyboard \circlearound{2} or physical buttons on the touchscreen \circlearound{3}.
In comparison, in Direct Control \circlearound{1} the vehicle is controlled via steering angle inputs and acceleration or braking via buttons on the steering wheel.

\subsection{Implementation}
\label{im}

\begin{figure}[htbp]
  \centering
  \includegraphics[width = 8.7cm]{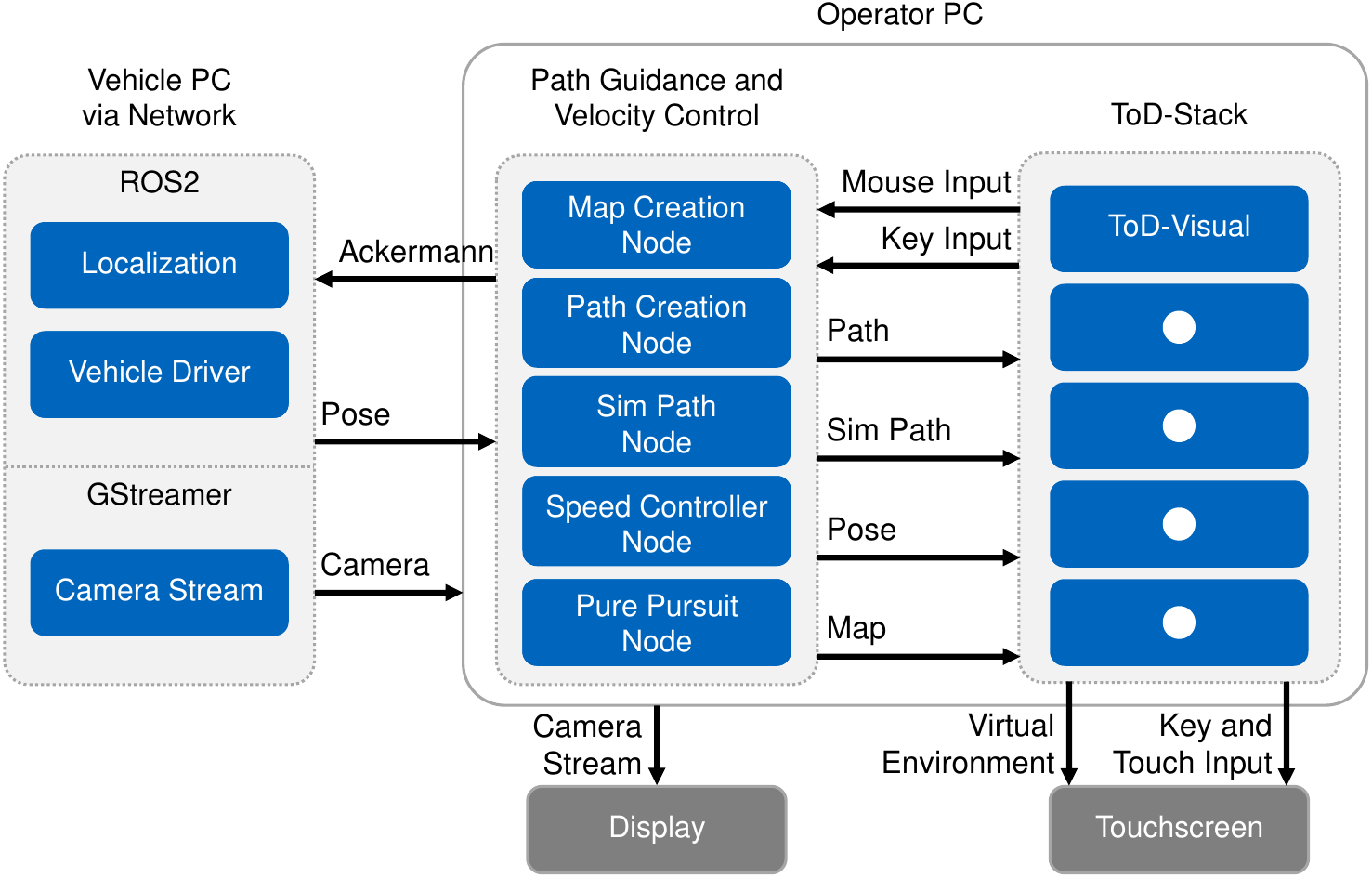}
  \caption{An overview of implemented modules and interfaces of the teleoperation system for separate path and velocity input. Arrows mark communication interfaces.}
  \label{fig:HMI_Overview}
\end{figure}

The software architecture is shown in Fig.~\ref{fig:HMI_Overview} and builds on the software stack for \ac{tod} from Schimpe et al. \cite{schimpe2022} converted to the ROS2 middleware. 
The \ac{tod}-Stack \cite{schimpe2022} contains several packages including the visualization.
Based on this, the package for path and velocity input comprises the stored map, the calculated path from the \ac{ro} input, the simulated actual path to be taken by the vehicle as well as the longitudinal and lateral commands for vehicle control.
These modules of the new package as well as the \ac{tod}-stack run on the operator side. 
On the other hand, localization, low-level actuator control, and video stream upload are performed on the vehicle PC.

For generating the virtual environment, the test track was scanned by a lidar sensor and the recognized objects were manually added to the map as blue boxes. Thus, the environment was premapped and no online object detection took place. 
The path generation utilizes the parametric spline interpolation from the open-source library developed by Kluge \cite{spline2021}. 
To simulate vehicle motion, a single-track model is used with parameters derived from the teleoperated vehicle.
The used test platform is an F1Tenth vehicle \cite{f1tenth2020} on a scale of 1:10 with a computing unit and sensors that enable the vehicle to drive itself and to be teleoperated via a wireless network.
A zed2 stereo camera from Stereolabs captures visual information about the vehicle environment, with the left camera image displayed on the \ac{ro} screen.

\subsection{Evaluation}

Following the implementation and realization of the three different prototype concepts an expert study was conducted. 
Due to the challenging teleoperation task, experts with previous experience in remote control were selected to participate in the study.
The aim was to ensure that the workload did not increase due to a higher learning effort.
Since, the number of people with teleoperation experience is limited and according to Nielsen \cite{Nielsen1994}, even a small sample size can reveal most usability problems of a given interface, seven teleoperation experts with academic background were part of the study.

For the data collection, a mixed-method approach was applied using a combination of the evidence from quantitative data and the exploratory insights offered by qualitative surveying \cite{almalki2016}.
Therefore, the standardized \ac{sus} \cite{Brooke1995}, NASA-TLX \cite{Hart2006} scores, task completion time and number of collisions were collected.
In addition, a ranking of the tested \ac{hmi} variants and feedback on how to improve the prototypes was requested.

During the study, the experts tested the three \ac{hmi} variants (Fig.~\ref{HMI_Konzepte}) on an approximately 30 meter long test track that had been shown in advance. 
The order in which the \ac{hmi} variants were tested was randomized between the test subjects.
After the participants had familiarized themselves with the system, they were requested to drive a lap on the test track, which was marked out with foam cubes in a test hall.
For the survey, the vehicle's maximum speed was set to \SI{0.35}{\meter/\second}, which corresponds to an upscaled walking speed of \SI{12.6}{\km/\hour} and allows the miniature vehicle to drive without jerking.
After each test drive, a questionnaire was filled out while the next \ac{hmi} setup was prepared by the test supervisor.


\section{Results}

This section describes the procedure for statistically analyzing the measured data and the results.
As the study analyzed a linked or dependent sample of seven experts (N=7) who tested all three different \ac{ui} variants, a Friedman ANOVA or test
was conducted. The Friedman test is based on the data ranking and shows whether there is a difference in the three design variants (\ac{sw}, \ac{mk}, \ac{ts}) for the measured data.
Differences were recognized in all objective measurements apart from the number of \textit{Collisions}. No \textit{Collisions} were detected during the entire course of the study.
Table \ref{tab2} shows that all $\chi^2$ values are above the critical value of 5.99 (\textit{df}\;=\;2 and \textit{$\alpha$}\;=\;0.05).

\begin{table}[htbp]
\vspace{-2mm}
\caption{Results of the Friedman Tests}
\begin{center}
\vspace{-0.1cm}
\begin{tabular}{|l|c|c|}
\cline{2-3}
\multicolumn{1}{l|}{N = 7 \qquad df\;=\;2} & $\chi$$^2$ & p-value\\
\hline
Task Completion Time & 11.14 & 0.00381\\
System Usability Scale & 6.46 & 0.03953\\
NASA Task Load Index & 6.00 & 0.04979\\
\hline
\end{tabular}
\label{tab2}
\vspace{-1mm}
\end{center}
\end{table}

\begin{figure}[!b]
  \vspace{-0.2cm}
  \centering
  \includegraphics[width = 8.5cm, scale=.9]{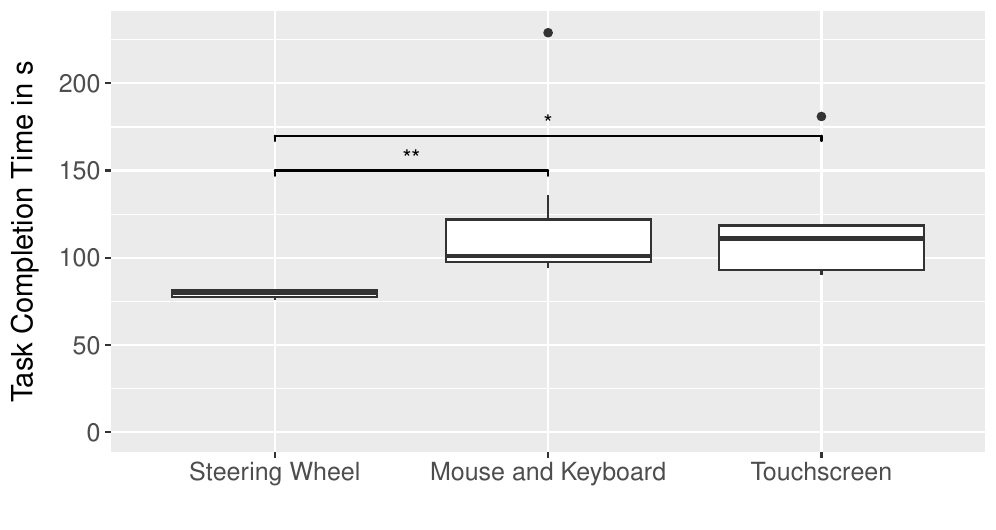}
  \vspace{-0.5cm}
  \caption{The boxplots show that the parallel input of path and velocity via steering wheel (\textit{Mdn} = $\SI{80}{\second}$) was significantly faster than the separate input via mouse and keyboard (\textit{Mdn} = $\SI{101}{\second}$) or touchscreen (\textit{Mdn} = $\SI{111}{\second}$).}
  \label{fig:TCT} 
\end{figure}

To reveal significant differences between the compared variants, the Friedman Multiple Comparisons Test was carried out.
The compared pairs that showed significance in their results were then analyzed pairwise using another post-hoc test, the Wilcoxon rank sum test. 
With regard to the \textit{Task Completion Time}, the evaluation showed that there is a significant difference between the parallel and sequential input of path and velocity (\ac{sw} and \ac{mk} with \textit{p}\;=\;$0.01563$
and \textit{r}\;=\;$0.896$; \ac{sw} and \ac{ts} with \textit{p}\;=\;$0.02225$ and \textit{r}\;=\;$0.896$) but not in the variants of the input with \ac{mk} and \ac{ts}, see Fig.~\ref{fig:TCT}.
Since both correlation coefficients \textit{r} are greater than 0.5, there is a strong effect and therefore a large difference in both cases according to Cohen \cite{Cohen1992}. This means that execution using conventional Direct Control concepts is significantly faster than separate path and velocity specification.

\begin{figure}
  \vspace{-0.2cm}
  \centering
  \includegraphics[width = 8.5cm]{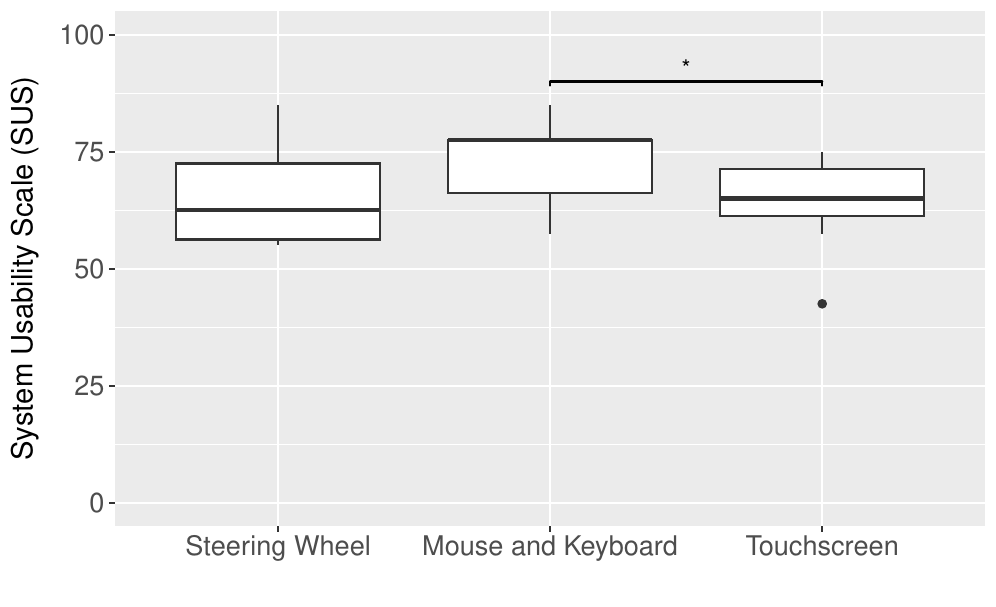}
  \vspace{-0.5cm}
  \caption{The usability evaluation revealed a significant difference between the sequential path and velocity input via mouse and keyboard (\textit{Mdn} = $77.5$) and touchscreen (\textit{Mdn} = $65.0$).}
  \label{fig:SUS} 
\end{figure}

In contrast, the evaluation of \textit{Usability} using the \ac{sus} only showed a significant difference in the variants of the path specification with different input devices with \textit{p}\;=\;0.02154 and \textit{r}\;=\;0.9. This difference also has a high effect size due to the \textit{r}-value above 0.5.
According to Bangor et al. \cite{Bangor2009}, \textit{Usability} is acceptable from an \ac{sus} score of 70 or higher. On average, only the version with separate path and velocity specification with \ac{mk} achieves this value, see Fig.~\ref{fig:SUS}. The variants with \ac{sw} and \ac{ts} are slightly below the acceptable threshold and can be categorized as marginally high.

\begin{figure}
  \vspace{-0.2cm}
  \centering
  \includegraphics[width = 8.5cm]{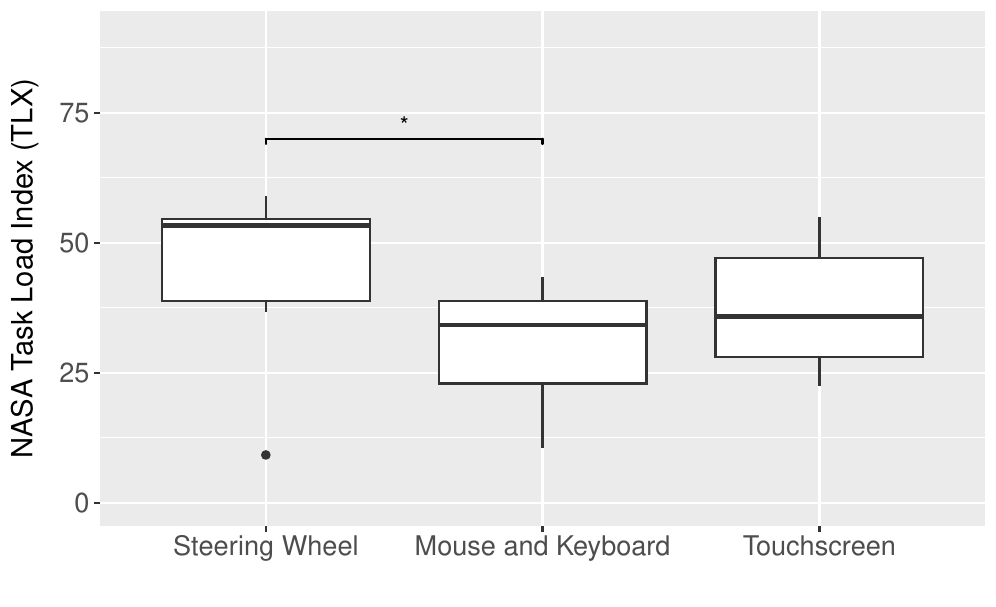}
  \vspace{-0.5cm}
  \caption{The workload for input via mouse and keyboard (\textit{Mdn} = $34.17$) is significantly lower than input via the steering wheel (\textit{Mdn} = $53.33$). In general, the workload for sequential path and speed input is lower.}
  \label{fig:NASA}
  \vspace{-3mm}
\end{figure}

The \textit{Workload} visualized in Fig.~\ref{fig:NASA} is only significantly higher in Direct Control using \ac{sw} compared to separate path and velocity specification using \ac{mk} with \textit{p}\;=\;0.2033. The effect size is \textit{r}\;=\;0.514 and, according to Cohan \cite{Cohen1992}, corresponds to a strong effect. 
When analyzing the individual scales of the NASA-TLX, it is particularly noticeable that the parallel input of path and velocity with \ac{sw} is not ahead in any of the categories, see Table \ref{tab3}.
In contrast, the separate input via \ac{mk} excels in low physical demand and frustration.
Additionally, the \ac{ts} beats the other variants in terms of temporal demand and the subjects' self-assessed performance.

\begin{table}[htbp]
\vspace{-3mm}
\caption{NASA-TLX Scales, Mean Values}
\begin{center}
\vspace{-0,2cm}
\begin{tabular}{|l|l|l|l|}
\cline{2-4}
\multicolumn{1}{l|}{} & \hspace{-1mm}Steering Wheel\hspace*{-0.5mm} & \hspace{-1mm}Mouse\&Keyboard\hspace*{-0.5mm} & \hspace{-1mm}Touchscreen\hspace*{-0.5mm}\\
\hline
\hspace{-1mm}Mental Demand & \hspace{-1mm}52.9 & \hspace{-1mm}35.7 & \hspace{-1mm}37.1\\
\hspace{-1mm}Physical Demand & \hspace{-1mm}22.9 & \hspace{-1mm}5.3 & \hspace{-1mm}37.9\\
\hspace{-1mm}Temporal Demand  & \hspace{-1mm}46.4 & \hspace{-1mm}37.3 & \hspace{-1mm}19.3\\
\hspace{-1mm}Effort & \hspace{-1mm}46.4 & \hspace{-1mm}25.0 & \hspace{-1mm}29.3\\
\hspace{-1mm}Performance & \hspace{-1mm}72.7 & \hspace{-1mm}69.3 & \hspace{-1mm}87.7\\
\hspace{-1mm}Frustration & \hspace{-1mm}22.9 & \hspace{-1mm}8.6 & \hspace{-1mm}14.7\\
\hline
\end{tabular}
\label{tab3}
\vspace{-1mm}
\end{center}
\end{table}

In the general ranking, the test subjects favored separate path and velocity input using \ac{mk}. 
The \ac{ts} variant was most frequently voted into second place. The conventional control via \ac{sw} was least favored.

The qualitative responses of the study participants were divided into the following categories in the analysis:
Feedback on the interaction concept, input device, map, camera image, viewing direction and additional \ac{ui} elements.

Favorable remarks regarding the interaction concept of parallel input of path and velocity \circlearound{1} highlighted the ability to cut curves, execute interactions quickly without premeditated planning, and a participant expressed a heightened sense of responsiveness to unforeseen events.
However, the control was reported to feel indirect and network latencies make the execution more demanding. The velocity input via the buttons on the \ac{sw} was rated particularly poor with the inaccessibility of the buttons leading to mistakes when cornering.
Against this, the map was assessed as being helpful when cornering, as areas outside the camera image could be viewed. Conversely, one subject stated that the map was irritating because the displayed representation of the surroundings did not correspond exactly to reality. As a result, the participant navigated in curves with larger radius than required.
However, two test subjects stated that the map information was hardly or not used at all during parallel input via the steering wheel because the focus was on the camera image. 
Meanwhile, only one participant noted poor image quality and that the image was slightly offset to the left and distracting in right-hand bends. 
In addition, the driving lane projection and speed information were requested.

Regarding the implementation of the interaction concept with separate path and velocity input in \circlearound{2} and \circlearound{3}, it was noted that the input of the trajectory was intuitive, regardless of the input device. The adaptation of points and the simulated path calculated by the vehicle were also mentioned positively.
However, one subject criticized the curve radius setting. Another criticism was that the line generated by the points was not intuitive, as the vehicle tended to cut the curves.
Opinions were divided on planning waypoints and monitoring at the same time, as it would not be possible to concentrate on both.
Moreover, test subjects criticized the orientation of the map, as it rotated in line with the vehicle's movements. Also, the area to be driven through could not be seen at a glance on the map. 
Nevertheless, the map was described as particularly useful when the video quality was poor.
Accordingly, the participants wanted a fixed map that can be moved and zoomed by the \ac{ro}.
Considering the different input devices, it was mentioned that the keyboard would be more intuitive to set the velocity but the stylus and tablet were found to be more comfortable by another test person.
In terms of screen layout, it was noted that in the \ac{ts} variant, the focus is on the map rather than the camera image. To avoid switching views, the information should be displayed on one screen or as an overlay where possible.
A warning system or collision information was requested as a further overlay to the camera image.


\section{Discussion and Design Suggestions}
\label{discussion}
 
In the following, the findings of the expert study are discussed and design guidelines are derived.

\subsection{Interpretation of Results}

In general, control using parallel path and velocity input is significantly faster than sequential execution and seems to react more quickly to unpredictable events. 
However, the test subjects preferred the sequential path and velocity input and rated the required workload lower. 
Input via \ac{mk} was particularly favored and rated as having the lowest workload but highest usability.
This confirms the initial assumption that the remote driving task is simplified for the \ac{ro} by sequential path and velocity specification.

It is therefore advisable to apply the concept of separate path and velocity specification in non-time-critical situations so that the \ac{ro} can complete the remote driving task as error-free as possible.
However, if the \ac{ro} needs to perform the task as efficiently as possible, i.e. quickly, conventional Direct Control via \ac{sw} is recommended. Due to the high workload using the \ac{sw}, appropriate \ac{ro} training is urgently required.

Regarding the input device for the separate guidance, it is surprising that \ac{mk} is preferred to the \ac{ts}, although the \ac{ts} would be the natural and direct input on the screen.
However, it can be assumed that the experts prefer this learned and therefore intuitive input due to their daily use of \ac{mk}.

\subsection{Design Guidelines}

In order to realize the separate input of path and velocity as user-friendly as possible, the following design guidelines are based on the feedback and findings of the study.

\begin{itemize}

  \item Both the camera image and the map should be displayed on one screen in order to keep all information in view and avoid distracting the \ac{ro} gaze, taking into account the human ergonomics.
  \item In general, the map has proved to be helpful as an abstract representation of the surroundings. However, it should be noted that map information may only be displayed in the visible sensor area of the vehicle camera or lidar if no premapped data or information from other vehicles or the infrastructure is available.
  \item The \ac{ro} should be able to move and zoom the map to view the whole driving area. The perspective on the map should be fixed and not following the vehicle movement.
  \item Velocity control buttons must be assigned intuitively and should not be on input devices that need to be moved. 
  \item The entered path should be adaptable. However, entering new waypoints at the end of the path while driving is considered critically. If necessary, the path length should be fixed as soon as the vehicle starts moving.
  \item Even if not everyone used the simulated vehicle path function, it should be implemented so that the user gains an intuitive understanding of the vehicle's behavior.
  \item The camera image should display the driving lane projection in order to be able to evaluate the vehicle behavior in advance.
  \item The \ac{gui} should include additional information on vehicle status and vehicle movement, e.g., the speed.
  \item In addition, the teleoperation system should have a collision avoidance or warning function to help estimate distances.
\end{itemize}


\section{Limitations}

The results of the expert study are based on certain limitations, which are referred to in this section and must be taken into account when using the findings.

The study has so far only been carried out with a small sample size. 
Furthermore, as discussed in Section \ref{discussion}, the experts may have favored \ac{mk} due to familiarity from their daily work. To assess the intuitiveness of the input device, the next study should involve non-experts.

It should be noted that the arrangement of the screens could also have had an influence on the evaluation of the individual variants in the study.
A further limitation is that the participants knew the test track. Although the test subjects operated with their backs to the track, there was no separation from the \ac{ro} workstation. 
Another limiting factor is that a miniature vehicle with only one camera stream was operated which restricted the field of view. 
Although the miniature vehicle can replicate the driving dynamic characteristics of a real vehicle in general, it cannot reproduce them fully.
However, the developed concept can be transferred to the interaction with a real vehicle via the same interfaces. The generation of the spline would also remain the same, whereby the sense of responsibility and the risk assessment of the test subjects would probably differ.

To test the \ac{hmi}, the vehicle was teleoperated along a relatively wide route in a static setting without unexpected disturbances, difficult narrow sections or other challenging elements apart from right and left turns.
The lack of collisions in this study is thus to be expected.
The study should be repeated in more difficult test scenarios so that monitoring and collision avoidance by the \ac{ro} are necessary.

The environment was represented by blue boxes in the map. As a result, the actual shape of the environment was lost. According to the observations of the test supervisor, the corners were cut less in the separate path and velocity specification, as the points were specified in the map, whereby the focus was on the camera image for the conventional Direct Control.
A test in a real road environment is pending.


\section{Conclusion}

The comparison of Human-Machine-Interfaces (\ac{hmi}) for teleoperation of vehicles demonstrated a preference among participants for the sequential input of path and velocity through the clicking of waypoints.
Mouse and keyboard (M\&K) emerged as the favorite input device.
Subsequently, there is a need to investigate the applicability of this approach in specific scenarios, particularly in challenging situations.
Further investigations are required to enhance the graphical display.
The study also revealed the connection between the individual teleoperation system components, shown in Fig.~\ref{fig:teleopSystem}.
It was observed that the User Interface (UI) can be improved through Safety Concepts such as collision avoidance by the vehicle.
Human workload, and thus potential errors, could be also reduced by the vehicle taking over the velocity specification through an alternative Teleoperation Concept, such as Waypoint Guidance.
This suggests new research areas in teleoperation that are largely unexplored so far.





\section*{ACKNOWLEDGMENT}

As first author, Maria-Magdalena Wolf initiated the research and carried out the literature survey and data analyses presented in this work. Furthermore, she instructed the system’s conceptualization, implementation and evaluation by setting the necessary framework conditions.
Richard Taupitz, as second author, was responsible for the system's design, implementation and execution of the study including the data collection and contributed to the literature research and writing process.
Frank Diermeyer made essential contributions to the conception of the research project on road vehicle teleoperation and revised the paper critically for important
intellectual content. 
He gave final approval for the version to be published and agrees to all aspects of the work. 
As a guarantor, he accepts responsibility for the overall integrity of the paper. 
Last but not least, we would also like to thank our colleague Tobias Kerbl for his introduction to the implementation and ongoing support.
The research was funded by the Federal Ministry of Economic Affairs and Climate Action of Germany (BMWK) within the project SAFESTREAM (FKZ 01ME21007B).

\bibliographystyle{IEEEtran}
\bibliography{bibliography}

\end{document}